\documentclass[runningheads]{llncs}
\usepackage{graphicx}
\usepackage[pagebackref,breaklinks,colorlinks]{hyperref}
\usepackage{tikz}
\usepackage{comment}
\usepackage{amsmath,amssymb} 
\usepackage{xcolor}
\usepackage{makecell}
\usepackage[symbol]{footmisc}
\renewcommand{\thefootnote}{\fnsymbol{footnote}}
\usepackage{orcidlink}
\usepackage[accsupp]{axessibility}  
\definecolor{darkgreen}{rgb}{0,0.7,0}

\newcommand{\degree}[1]{°\ }
\newcommand{\zfl}[1]{{\color{black}#1}}

\newcommand{\csb}[1]{{\color{black}#1}}
\usepackage{lipsum}

\newcommand\blfootnote[1]{%
  \begingroup
  \renewcommand\thefootnote{}\footnote{#1}%
  \addtocounter{footnote}{-1}%
  \endgroup
}
\begin{document}
\pagestyle{headings}
\mainmatter
\def\ECCVSubNumber{6182}  

\title{Deep 360° Optical Flow Estimation Based on Multi-Projection Fusion} 

\titlerunning{ECCV-22 submission ID \ECCVSubNumber} 
\authorrunning{ECCV-22 submission ID \ECCVSubNumber} 
\author{Anonymous ECCV submission}
\institute{Paper ID \ECCVSubNumber}

\titlerunning{Deep 360\degree\ Optical Flow Estimation}
%
\author{Yiheng Li\inst{1} \and
Connelly Barnes\inst{2} \and
Kun Huang\inst{1} \and
Fang-Lue Zhang\inst{1,*}
\orcidlink{0000-0002-8728-8726}
}
\authorrunning{Y. Li et al.}
%
\institute{School of Engineering and Computer Science, Victoria University of Wellington
\email{\{Yiheng.Li,Kun.Huang,Fanglue.Zhang\}@vuw.ac.nz}\and
Adobe Research, Seattle, US\\
\email{ConnellyBarnes@gmail.com}}
\maketitle
\blfootnote{* Corresponding author}
\begin{abstract}
Optical flow computation is essential in the early stages of the video processing pipeline. This paper focuses on a less explored problem in this area, the 360\degree\ optical flow estimation using deep neural networks to support increasingly popular VR applications. To address the distortions of panoramic representations when applying convolutional neural networks, we propose a novel multi-projection fusion framework that fuses the optical flow predicted by the models trained using different projection methods. It learns to combine the complementary information in the optical flow results under different projections. We also build the first large-scale panoramic optical flow dataset to support the training of neural networks and the evaluation of panoramic optical flow estimation methods. The experimental results on our dataset demonstrate that our method outperforms the existing methods and other alternative deep networks that were developed for processing 360\degree\ content. 
\keywords{Optical Flow, 360\degree\ Video, Spherical Image Projection}
\end{abstract}

\section{Introduction}

If we take a sphere representing all viewing directions for 360\degree\  images and remove one point from it, then there is no projection of the resulting surface to a 2D plane that is both an equal area map and angle-preserving (conformal). This is because the sphere is not developable \cite{GeometryBook}, as is well-known from cartography. This mathematical fact led to classical research into mesh parameterizations such as \cite{nadeem2016spherical,poranne2017autocuts} that \csb{ trade off between competing goals} like area and angle-preservation when mapping textures on spherical surfaces. In this research, we were initially motivated to discover good projections for optical flow learning for 360\degree\ videos, which remains an interesting and open question in this field. We investigated three different projections from the sphere to the plane: cylindrical and cube-map, which are conformal within each chart, and equirectangular, which is neither area- nor angle-preserving. However, to our surprise, our experiments did \emph{not} seem to indicate that any of our projections is always better than the others, but rather that these projections make \emph{complex tradeoffs} in optical flow accuracy that appear to depend jointly on the input image and the mathematical properties of each projection. 

Since empirically no single projection appears to always be best, we therefore propose to learn optical flow using multiple projections: \csb{we} learn to fuse the complementary information into a single final optical flow on the sphere. We find that such learned fusion of multi-projection optical flows consistently outperforms the single-projection optical flows that were fused. We thus propose that multi-projection optical flow with fusion can be an important principle for 360-degree optical flow, even though more remains to be discovered in future work about e.g. the desired mathematical properties of good projections. Since there are no existing large-scale datasets with ground truth panoramic optical flow to learn the full field-of-view (FOV) features, we also build the first 360\degree\ optical flow dataset that contains various scenes and dynamic objects, including around 6000 panoramic frames for training our deep models. 

Besides the above issue due to the mathematical properties of \csb{projections for} 360\degree\ optical flow, there are common challenges with narrow FoV videos such as robustness to outliers, illumination changes, and possible large displacements, which have been well addressed in existing non-360\degree\ deep optical flow deep models~\cite{ren2017unsupervised,DFIB15,IMKDB17,Sun2018PWC-Net,DBLP:journals/corr/abs-2003-12039}. Although such motion models are formulated on 2D regular grids with convolutions that do not inherently deform according to distortion or area changes in the projection, we can still utilize them as strong backbone models to generate accurate complementary regions for different projections. To make the maximum use of their pre-trained models that can handle complex dynamic objects, we did not use spherical kernels~\cite{coors2018spherenet} to modify their convolutional layers, as we found the modified receptive field of such kernels led to worse results when finetuning the backbone models using our datasets. 

In summary, this paper makes the following contributions: 1. \zfl{We introduce a fusion framework for learning 360\degree\ optical flow}, which learns to fuse the complementary optical flow prediction results generated using different projections; 2. We apply common map projection techniques to make two spherical image projection methods, namely tri-cylinder projection and cube padding projection that are complementary to equirectangular projection in terms of distorted \csb{regions} and performance under fusion; 3. We build a novel 360\degree\ optical flow dataset with ground truth optical flow data with various camera paths and dynamic objects.

    




\section{Related Work}
This section reviews prior work in the relevant domains of optical flow estimation, datasets for training deep models, and 360\degree\ image and video processing.  

\noindent\textbf{Optical Flow Estimation}
Optical flow shows the motion of objects in a video by a dense vector map pointing to the corresponding pixels between frames. The traditional methods mostly rely on feature constancy and local spatial coherence to solve the problem in an optimization framework~\cite{10.5555/1623264.1623280,horn1981determining,sun2010secrets,Liu_2014_CVPR,liu2016meshflow}. In the era of end-to-end deep learning, the advances in optical flow estimation are driven by deep models ~\cite{Opt2019Survey}. The work of FlowNet~\cite{DFIB15} built a solid foundation for optical flow prediction using deep networks. They use FlowNetSimple to learn optical flow directly from adjacent frame pairs and use FlowNetCorrelation to learn the correlation between the patch-level features of the pair of input frames. FlowNet2~\cite{IMKDB17} presents some improvements based on FlowNet by proposing a stacked architecture and training schemes that benefit the convergence of the training process. PWC-Net~\cite{Sun2018PWC-Net} proposed to use Pyramid, Warping, and Cost volume to simplify the network structure and achieve high performance in optical flow estimation. RAFT~\cite{DBLP:journals/corr/abs-2003-12039} uses Recurrent All-Pairs Field Transforms that apply all-pairs correlation information into the neural network.
In contrast, the above models only use nearby patches for correlation information for saving memory. Some other works focus on unsupervised learning schemes~\cite{meister2018unflow,wang2018occlusion,liu2019selflow,liu2020Analogy}, \zfl{fusing flows from multiple frames}~\cite{ren2019fusion}, and joint learning of optical flow and occlusions~\cite{JointOptOcc}, aiming to train networks with the objective from view synthesis~\cite{yin2018geonet}. Despite promising results, generalizing these methods beyond narrow FOV videos remains challenging. 

\noindent\textbf{Optical Flow Dataset}
Current optical flow prediction methods treat optical flow estimation as a supervised learning task that requires a considerable amount of data for training. As the real-world optical flow data is hard to access, most datasets are synthetic and rendered by Blender, while few of them are collected by camera and \csb{LIDAR}. 
Sintel~\cite{Butler:ECCV:2012} is derived from a 3D movie to provide thousands of image pairs with dense ground truth optical flow. Wulff et al.~~\cite{Wulff:ECCVws:2012} further extend it to two versions, Clean and Final, which contains RGB albedo and visual effects such as motion blur and fog, respectively. FlyingChairs~\cite{DFIB15} is generated by a set of 3D chairs models and applies the random 2D affine transformation to the scene. The idea was then expanded to generate FlyingThings3D~\cite{DBLP:journals/corr/MayerIHFCDB15} from the assets published in Stanford's ShapeNet, where random motions were applied on the dynamic objects. KITTI~\cite{Geiger2012CVPR}~\cite{Menze2015CVPR} dataset was collected from real-world by cameras and LIDAR installed on a car. However, due to the device limitation, their optical flow is sparse, especially for the natural background where the \csb{LIDAR} could not provide valid depth information. To address the sparsity issue of the KITTI dataset, Mayer et al. released a dataset called Driving~\cite{DBLP:journals/corr/MayerIHFCDB15}, which renders a naturalistic city from the perspective of a driving car to generate dense optical flow maps.
Moreover, for the convenience of nonrigid and softly articulated motion analysis, Monkaa~\cite{DBLP:journals/corr/MayerIHFCDB15} is designed to resemble Sintel using cartoon animations. A small scale dataset of 360\degree\ optical flow was proposed recently by Yuan and Richardt~\cite{yuan2021360}, which can only be used for evaluation due to its limited size. To support the training of 360\degree\ optical flow estimation networks, a new dataset is needed in this domain.

\noindent\textbf{360\degree\ Image and Video Processing}
360\degree\ images and videos have been attracting increasing attention from computer vision and graphics researchers. Due to the special representation of 360\degree\ images, dedicated methods were developed for the analysis and processing tasks, such as depth estimation~\cite{wang2020360sd}, saliency detection~\cite{monroy2018salnet360,zhang2018saliency}, edit propagation~\cite{zhang2021efficient}, and video stabilization~\cite{kopf2016360,tang2019joint}. With the recent development of immersive technologies, researchers have been working on the specific problems of panoramic video-based applications, such as automatic 360\degree\ navigation/piloting ~\cite{360Nav,Deep360pilot}, 360\degree\ video assessment~\cite{li360Assess}, and immersive video editing~\cite{360VRediting}. The lack of reliable temporal correspondences that are globally consistent hinders many potential applications of 360\degree\ video generation. Our method is proposed to fill that gap.  

The majority of current convolutional networks were developed for coping with \csb{perspective projection onto a planar camera sensor}. Researchers proposed several solutions to adopt convolutions on spherical images to facilitate the feature extraction and interpretation by the deep networks. A representative method, namely SphereNet~\cite{coors2018spherenet,SphereNetTMM}, changes the perceptual field of their convolutional kernels according to the latitude on the equirectangular domain. However, changing the kernels of 2D optical flow networks to spherical kernels may degrade the effect of the pre-trained models, as shown in our experiments.
The kernel transform technique~\cite{ktnpaper} was also considered in this field to let the convolutional layers learn how to transform the spherical kernels to the pre-trained weights on regular kernels that were initially trained by perspective images. However, the current optical flow networks all have a large number of layers, causing an \csb{impractical} computation and memory cost to learn to transform all the layers. Projecting the spherical images to cube maps can avoid the distortion issues when utilizing deep neural networks, as in 360\degree\ saliency detection~\cite{cheng2018cubeSal}. However, after testing with cubemap-based projection, we found the inconsistency of the predicted results along the cube face boundaries can not be trivially addressed, which also inspires us that using any single projections would meet more or fewer issues in certain regions. We thus propose to fuse optical flow results predicted using different projections \csb{to utilize} their complementary information.


\section{Dataset Generation}
Due to the lack of ground truth data~\cite{bhandari2021revisiting} of 360\degree\ optical flow, existing optical flow estimation networks could not be appropriately trained for 360\degree\ videos. We thus build a dataset with ground truth panoramic optical flow maps by rendering synthetic 360\degree\ videos of dynamic 3D  scenes.

\subsection{Optical Flow Generation}
Two assumptions are made when generating panoramic optical flows when rendering a 3D scene, which are commonly followed by other optical flow datasets~\cite{DBLP:journals/corr/MayerIHFCDB15}. Firstly, all the visible points that appear on a frame can ideally be seen on the next frame. Secondly, the geometry of rendered objects is not deformed. Denote a pixel position as $p$, its position on the corresponding 3D object $O$~\cite{sterzentsenko2019denoising} can be calculated using its model transformation matrix $M$ that represents the object motion and the camera projection matrix $P$ by $O = M^{-1}P^{-1}I$. For the successive frame, assume $M^\prime$ and $P^\prime$ represent its model matrix and camera matrix, respectively, the corresponding image pixel positions for the point $O$ on the 3D object can be obtained by $I^\prime = P^\prime M^\prime O $. In equirectangular projection, optical flow needs to be transformed into spherical coordinates from a 3D point $(x,y,z)^T$ on the unit sphere in the camera space by:
\begin{equation}
\theta^\prime = \mathrm{atan}\left(\frac{z^\prime}{x^\prime}\right), \ \ 
\phi^\prime = \mathrm{atan}\left(\frac{y^\prime}{\sqrt{(x^\prime)^{2} (z^\prime)^{2}}}\right)
\end{equation}

From this, we can obtain the optical flow vectors $(\bigtriangleup\theta, \bigtriangleup\phi)$ in spherical coordinates. We can easily convert and store these in an equirectangular projection-based representation. More details are included in the supplementary materials.


\subsection{3D Scenes}
We built our panoramic RGB and optical flow frames using two types of scenes: 

\noindent\textbf{City Scene}
We designed the City scene to resemble the data with similar properties as KITTI~\cite{Geiger2012CVPR} and Driving~\cite{DBLP:journals/corr/MayerIHFCDB15}. The city's layout is automatically procedurally generated, and over 300 vehicles are randomly moving in the city. We enable collision detection for all the objects in the scene to ensure that it can avoid the situations of object-object and object-camera interpenetration. For the pre-training and preliminary experiments, we built two subsets named City1000UR and City100UR, which contain 1011 and 102 up-right frames for fine-tuning and testing. We divide the whole 3D city into two parts, where half is used for generating training data and the other half for the validation and testing data. We then generated City2000, City200, and City100 datasets, where the camera is rotated within 45\degree\ , containing 2000, 217, and 138 frames, respectively. 

\noindent\textbf{Equirectangular FlyingThings Scene}
We designed the EquirectFlyingThings scene to resemble FlyingChairs~\cite{DFIB15} and FlyingThings3D~\cite{DBLP:journals/corr/MayerIHFCDB15}. In this scene, objects are randomly distributed around the camera. In our implementation, we choose car models from Stanford's ShapeNet dataset\cite{shapenet2015}, which contains complex mesh shapes and textures.
Compared to FlyingThings3D, the EquirectFlyingThings dataset contains more objects in the scene because the camera takes panoramic photos. We rendered 300 objects to fill in the content and scale them to a suitable size. We also apply random translation and rotation to those objects. Furthermore, we change the camera rotation direction every 20 frames to keep randomness without camera shake. We built EFT2000, EFT200, and EFT100 datasets, and each of them contains 2211, 199, and 99 image pairs.

For our final models, we take the City2000 and EFT2000 as training datasets, while the City200 and EFT200 are used for validation. The final testing is performed on the City100 and EFT100 datasets.


\section{Multi-Projection Fusion}
To maximally leverage existing narrow FOV pre-trained models while minimizing the impact of various distortions that occur in spherical projections, we propose to use different methods to project the spherical images to planar images and feed them to finetune the models. A fusion layer is finally built to combine the transformed optical flow results for the final 360\degree\  optical flow. 

We first perform preliminary experiments to select the best baseline model to finetune and predict the optical flow with equirectangular projections. We adopt the widely used models FlowNet2~\cite{IMKDB17} and PWC~\cite{Sun2018PWC-Net}, and the most recent state-of-the-art method RAFT \cite{DBLP:journals/corr/abs-2003-12039} to finetune on our 360\degree\  dataset using equirectangular projection. We found that PWC achieved the highest performance and the equirectangular representation is the best single projection for optical flow estimation. Because our experiments indicate that fused models always perform no worse (and in fact, better) than single projections, we choose PWC as our baseline model and use it to finetune the optical flow prediction model for each projection. 
In addition, we also modified all the PWC-Net's convolutional layers to Spherical convolutional kernels as in \cite{coors2018spherenet}. However, the experimental results show that the finetuning with spherical convolutions cannot achieve a good performance due to kernel shape changes from the original pre-trained PWC model. More results are demonstrated in later sections.


\begin{figure}[t]
    \centering
    \includegraphics[width=.95\textwidth]{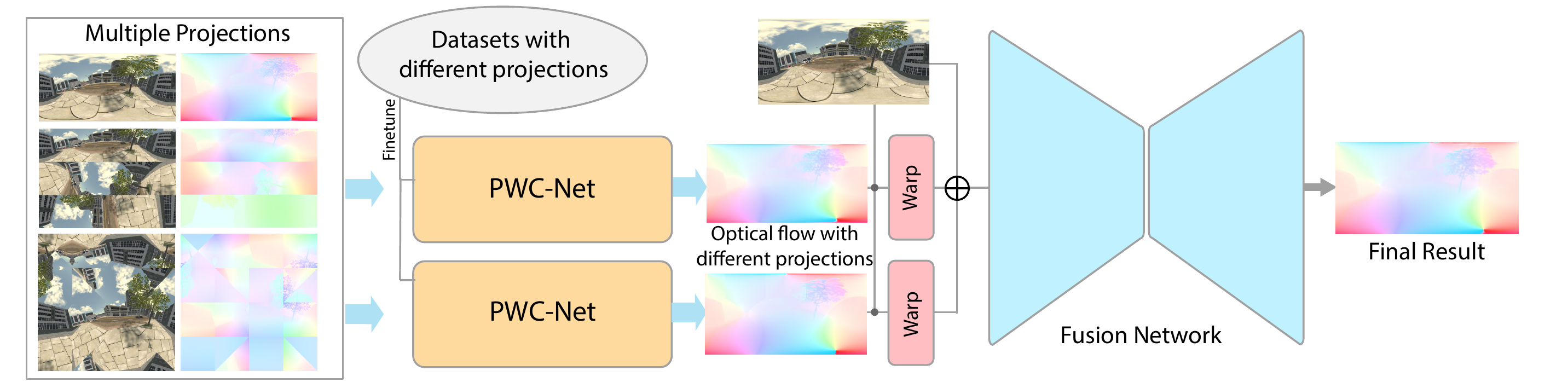}
    \caption{Our multi-projection fusion method for estimating 360\degree\ optical flow
    }
    \label{Pipeline}
\end{figure}

\subsection{Projections for Spherical Images}
Here, we investigate the performance of the deep models when feeding 2D images projected from the spherical domain using different \csb{kinds} of projections.
To utilize the pre-trained 2D deep models, ideally we would wish for the panoramic view to be an isometry to the original spherical surface, after making at least one cut. This ideal projection does not exist as the sphere is not developable, so distortion is necessary. In our experiments, we thus explore different mapping methods. It is also worth noting that we take the optical flow $C^{0}$ and $C^{1}$ continuity as another important consideration because we want the neural network to keep the integrity of moving objects. Some violations of these e.g. $C^{0}$ everywhere and $C^{1}$ almost everywhere may be acceptable given the tolerance of convolutional networks to various local transformations. 

\noindent\textbf{Equirectangular Projection}
For 360\degree\ video, equirectangular projection is the most popular way to map a spherical surface to a 2D rectilinear plane. It can keep both $C^0$ and $C^1$ continuity in the optical flow map except for the boundary areas, which can be addressed by the circular padding scheme of CNNs. However, it is not conformal and introduces large distortions due to excessively sampling the top and bottom of the sphere. 
When using equirectangular frame pairs and optical flow maps to train the baseline model, we further modify the loss terms to compensate for distortion by applying solid angle weights onto their pixel-wise loss terms. The solid angle weight is generated by projecting the spherical grid $\theta$ and $\phi$ to the corresponding $x$, $y$, and $z$ at the unit sphere presented in the Cartesian coordinate system.
The pixel position we calculated is represented as the center pixel $P_{c}$. Then, we generate its adjacent right and downside pixel position $P_{r}$ and $P_{d}$. \csb{Next,} we calculate the area covered by each pixel $S$ by: 

\begin{equation}
P = (P_{r} - P_{c})\times (P_{d} - P_{c}), \ 
S = \sqrt{P_{x}^{2} + P_{y}^{2} + P_{z}^{2}}
\end{equation}

The covered area is used as the pixel-wise weight term in the loss function to reduce the effect caused by oversampling in training and testing.

\noindent\textbf{Tri-Cylindrical Projection}
Unlike equirectangular projection, cylindrical projections preserve angle but are not equal-area. We use the Mercator projection to transform our equirectangular images to cylindrical images by converting the coordinate $(\theta_{e}, \phi_{e})$ to cylindrical coordinate $(\theta_{c}, h_{c})$ by:

\begin{equation}
\theta_{e} = \theta_{c}, \ \ 
\phi_{e} = 2\mathrm{tan}^{-1}(e^{l}) - \frac{\pi}{2}
\end{equation}

Here, $l$ is a value related to latitude in cylindrical coordinate. The inverse conversion between the two coordinates can be done using the following formula:

\begin{equation}
\theta_{c} = \theta_{e}, \ \ 
h_{c} = \mathrm{ln}(\mathrm{tan}(\phi_{c}) + \mathrm{sec}(\phi_{c}))
\label{equ:inv_equi_cyl}
\end{equation}

\begin{figure}[t]
    \centering
    \includegraphics[width=\textwidth]{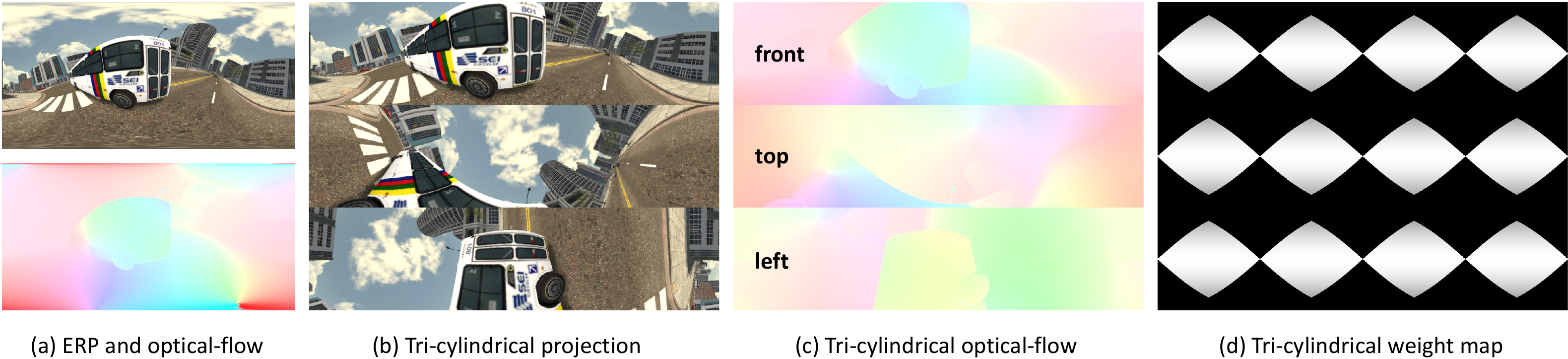}
    \caption{A tri-cylindrical projection example. Note that different projections provide similar appearance with perspective images for different regions. The weight map is masked after applying solid angle weights to ensure every pixel is just calculated once.}
    \label{Cylindrical_City_dataset_image}
\end{figure}

For a given cylindrical coordinate $(\theta_{c}, h_{c})$, we use its corresponding equirectangular pixel to find the target pixel using its flow vector and convert the target pixel to cylindrical coordinates by Eq.~\ref{equ:inv_equi_cyl} for obtaining the cylindrical optical flow. 
Since the Mercator projection generates a much larger area for the regions of high latitude, causing an even more serious distortion and oversampling problem, we just convert a limited vertical field of view to cylinders where the area change is less extreme (90 degrees in our experiment). To ensure that every part of the spherical surface has an equal contribution, we propose to stack three cylindrical projection images together, where each cylinder to project is aligned with one of the X, Y, and Z-axis, as shown in Fig. \ref{Cylindrical_City_dataset_image}.

To ensure a fair comparison, the solid angle weights for all the pixels are again generated using the same approach as in the equirectangular image. However, as we stack some pixels three times in the same image, we have to ensure every pixel is calculated only once.
We use the distances between the spherical pixels and the equator of each cylinder to decide which view to use for that point. Intuitively, if a pixel is closer to one equator of a cylinder, the distortion should be less, leading to a more accurate result. After applying such a mask, the normalized weight map for our tri-cylindrical projection is shown in Fig. ~\ref{Cylindrical_City_dataset_image}.

\noindent\textbf{Cube-Padding Projection}
Cube maps are another common representation of 360\degree\  images, with the advantages of using perspective images to represent a panoramic image. \csb{Within each face of the cube, the projection is angle-preserving but not area preserving,} and is suitable for the pre-trained model to be further finetuned. A drawback of the original cube map is that \csb{adjacency} information is lost if the zero values fill the space out of the cross formed by the six connected cube faces. To \csb{ensure} global $C^0$ continuity of the optical flow, i.e., connecting the outer boundary of the six faces to their neighboring faces in the cube, we repeatedly pad the faces to stitch the six faces into one picture, keeping the spatial continuity based on the original cross layout. 

\begin{figure}[t]
    \centering
    \includegraphics[width=\textwidth]{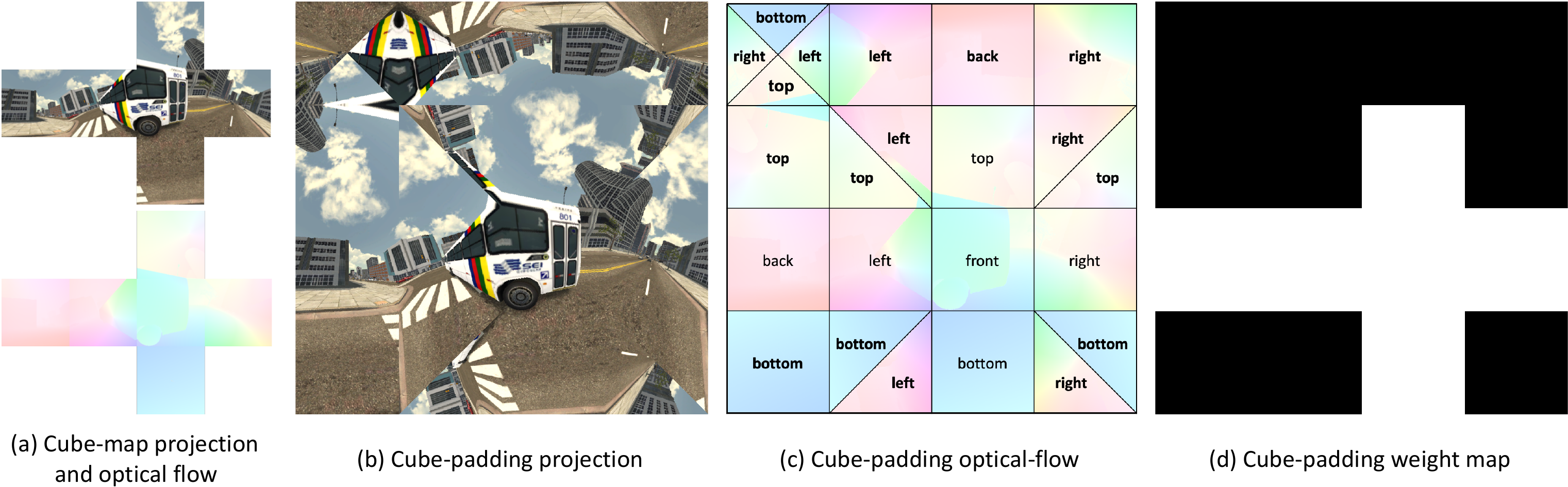}
    \caption{The 360° image and its optical flow projected by our cube-padding approach and the weight map applied on the loss functions.}
    \label{Cube-map-padding-RGB}
\end{figure}

The layout of the faces is shown in Fig. \ref{Cube-map-padding-RGB}, where some faces are rotated and trimmed to ensure the $C^0$ continuity along the boundaries. The drawback of this projection is that the optical flow can not keep $C^{1}$ continuity because the optical flow is projected onto image planes of different cameras, leading to the direction changes along the face boundaries. Thus, although using this kind of projection to finetune the baseline model can give a certain level of improvement, it does not outperform the equirectangular-based model in our later experiments. We use perspective projections to convert the spherical pixels to cube map faces for producing cube-padding projection results.


\subsection{Fusion from Multiple Projections}
\begin{figure}[t]
    \centering
    \includegraphics[width=\textwidth]{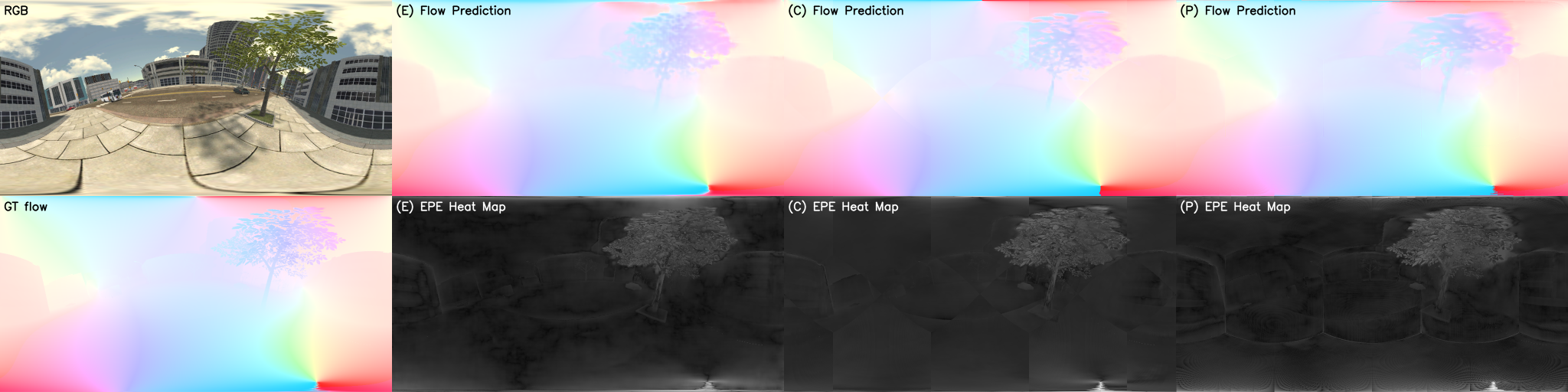}
    \caption{EPE maps for different projections: (E)-Equirectangular, (C)-Cylindrical, and (P)-Cube-Padding. Here, the EPE values are normalized for visualization.}
    \label{Heatmap}
\end{figure}

We observe that the models trained by different projections provide complementary results in different regions. The equirectangular projection model can better predict the optical flow of pixels around the equator. The cylindrical projection model has advantages at the top and bottom regions using the cylinders aligned with X and Z axis. The cube-padding projection can produce high-quality results besides all the boundary regions of the cube faces. Here we use error heatmaps to visualize different error distributions in Fig. \ref{Heatmap}. 
 The minimum EPE value presented here shows that the prediction can be improved if appropriately blended with these estimations.

\begin{figure}[t!]
    \centering
    \includegraphics[width=.9\textwidth]{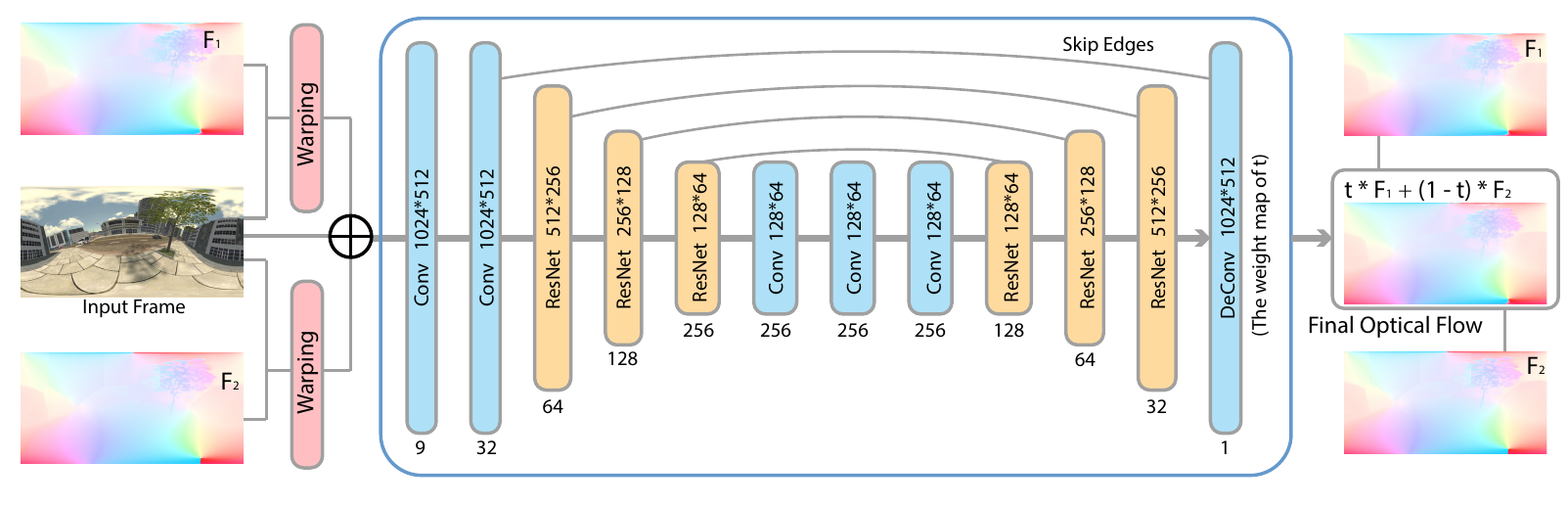}
    \caption{Fusion network structure. }
    \label{Fig:fusion}
\end{figure}
To utilize the advantages of different projections, we develop a fusion network to blend the result. The fusion network is a U-Net~\cite{ronneberger2015u} with ResNet layers~\cite{he2016deep} to learn a confidence parameter $t$. The model structure is shown in the following table. The ResEncoder and ResDecoder are based on the ResNet structure shown is the figure \ref{Fig:fusion}. In addition, we take the Mish~\cite{misra2019mish} as our activation function. The learnt confidence parameter is used to blend two models' predictions by: $
P_{final} = t * P_{a} + (1 - t) * P_{b}$, 
where a and b represent the predictions using different projection methods. More specifically, the prediction results are converted to equirectangular frame representations before fed to the network. Then the input frame will be concatenated with the warped frames by the two predicted optical flow maps before entering the convolutional layers. From the experiment results, fusing any two of the three models based on different projections can generate significantly better results than using an individual projection, showing the effectiveness of our fusion scheme for 360\degree\  optical flow estimation.

\begin{table}[t]
\centering
\begin{tabular}{|c|c|c|c|c|c|c|}
\hline
\multicolumn{1}{|c||}{Model} & \ \ \ FlowNet2 \ \  & \ \ PWC-Net\ \ &\ \ PWC-Sph\ \   & \ \  RAFT-12\ \  & \ \ RAFT-24\ \  &\ \ RAFT-48\ \ \\ \hline
\multicolumn{1}{|c||}{EPE}   & 4.15     & \textbf{3.17} & 3.80 & 5.12    & 4.93   & 4.72 \\ \hline

\end{tabular}
\caption{Performance on City100UR using the finetuned models.}
\label{table:comparison between FlowNet2, PWC and RAFT}
\end{table}
\section{Experiments}
We implement our method by utilizing state-of-the-art optical flow estimating models. In the preliminary experiments, we fine-tuned three state-of-the-art neural network structures: FlowNet2~\cite{IMKDB17}, PWC~\cite{Sun2018PWC-Net}, and RAFT~\cite{DBLP:journals/corr/abs-2003-12039} using our datasets. Our experiment found that the PWC structure with only supervising the final output layer can give us the best performance on our test datasets. Thus, we take the PWC structure for our backbone model. Our experiment result for different models is shown in Table \ref{table:comparison between FlowNet2, PWC and RAFT}. We trained PWC-Net using different loss terms and tried applying the normally used solid-angle weights \cite{zhao2021adaptive} to find out the best loss function as our backbone model. Finally, we \csb{chose} the top-layer loss without solid-angle weights as the best approach (see our supplementary materials). We also \csb{tried} modifying the PWC-Net architecture by changing their convolutional kernels to spherical kernels ~\cite{coors2018spherenet} to adapt to the equirectangular distortion. However, as the pre-trained weights are achieved using the regular convolutional kernel shape, the perceptual field changes led to unsatisfactory finetuning results as shown in Tab.~\ref{table:comparison between FlowNet2, PWC and RAFT}.



\subsection{Implementation}
For the data pre-processing and the projection conversion, we implement OpenCL programs to convert the equirectangular image into cylindrical projection and the planar padding format, respectively. Our model utilizes the PWC structure as the backbone, while the correlation layer is from the FlowNet2's implementation. \zfl{Our
model size is 9.4MB, and the parameter number is 2.46M.
All the components have an O(N) complexity. The running time of each component on an RTX3090 is reported in Tab.~\ref{tab:runtime}. Note that our implementations of the projection operations can satisfy real-time requirements.}

\begin{table}[b]
\centering
\resizebox{10cm}{!}{
\begin{tabular}{|c|c|c|c|c|}
\hline
Component  & PWC-Net  & Fusion Model & Projection-E2C & Projection-E2P \\ \hline
Time (ms) & 25.6 & 25.9   & 8.8 & 6.8  \\ \hline
\end{tabular}}
\caption{Running time of each component and projection.}
\label{tab:runtime}
\end{table}



\noindent\textbf{Data scheduling}
Panoramic videos contain irregular movements of objects and cameras, making the scene too complex to analyze. The experiment presented in FlowNet2 shows that training the network with simple features can help the learning process~\cite{IMKDB17}. Thus, we fine-tune models with City1000UR where the camera does not rotate about the x and z axis during data generation.

\noindent\textbf{Fusion network with extra parameters} We tested two versions of fusion layers. The first design takes the predictions $P_{e}$, $P_{c}$ and the source images $I_{s}$. The other version also utilizes two channels of the latitude and longitude $\theta$ and $\phi$ of each pixel to enrich the information the network takes. Due to the possible overfitting to the spatial position, the second model has a higher EPE (2.65 v.s. 2.61). Thus, we discard the spatial position channels in our final network.



\begin{figure}[t]
    \centering
    \includegraphics[width=.95\textwidth]{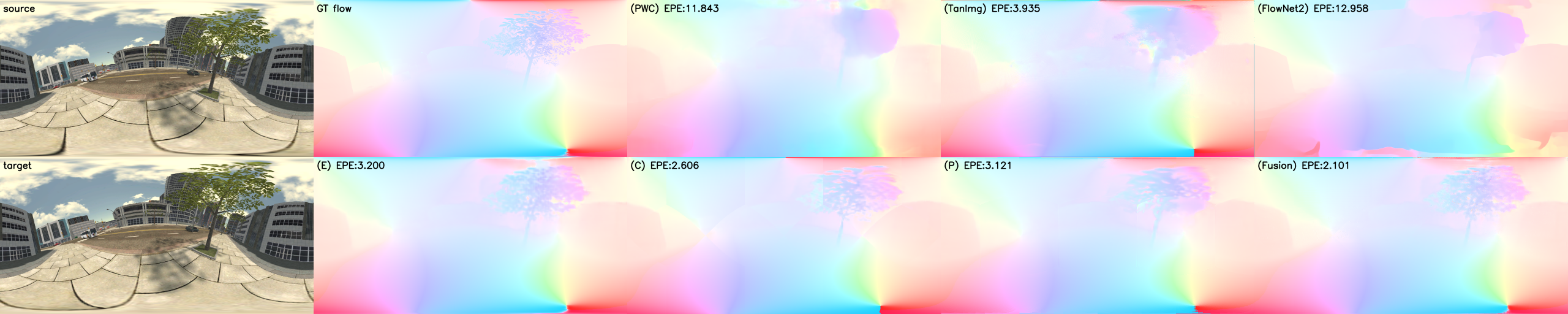}
    \includegraphics[width=.95\textwidth]{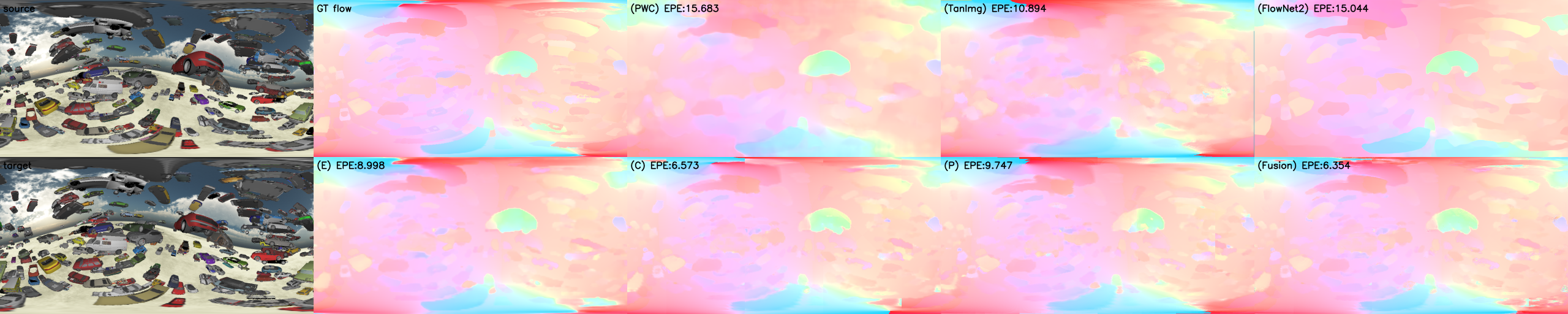}
    \caption{Optical flow estimation results using PWC, tangent image-based method (TanImg), FlowNet2, single projections, and our final fusion model.} 
    \label{fig:results}
\end{figure}

\begin{table}[t]
\centering
\begin{tabular}{|p{43pt}<{\centering}|p{43pt}<{\centering}|p{43pt}<{\centering}|p{43pt}<{\centering}|p{43pt}<{\centering}|p{43pt}<{\centering}|p{43pt}<{\centering}|}
\hline
\multicolumn{1}{|c||}{SD/EPE}   & Equi (E) & Pad (P) & Cyl (C) & C+P & E+P & E+C \\ \hline
\hline
\multicolumn{1}{|c||}{City100} & 0.90/2.41    & 0.92/3.32     & 1.34/2.64 &0.82/2.15 &\textbf{0.75}/1.95 & 0.83/\textbf{1.79} \\ \hline
\multicolumn{1}{|c||}{EFT100}  & 2.98/6.65    & 2.98/6.25     & 2.82/5.2  &2.77/5.06 &2.95/5.84 & \textbf{2.64}/\textbf{5.01} \\ \hline
\multicolumn{1}{|c||}{Average} & 1.94/4.91    & 1.95/5.68   &2.08/5.06 &1.80/3.60 &1.85/3.89 & \textbf{1.74/3.40} \\ \hline
\end{tabular}
\caption{\zfl{Spherical Distance (SD) / End point errors (EPE) using equirectangular (Equi), cube-padding (Pad), cylindrical (Cyl) projections and their fusion models.}}
\label{tab:compare_projs}
\end{table}
\subsection{Fusion from Different Projections}
\begin{figure}[t]
  \centering
\includegraphics[width=.95\linewidth]{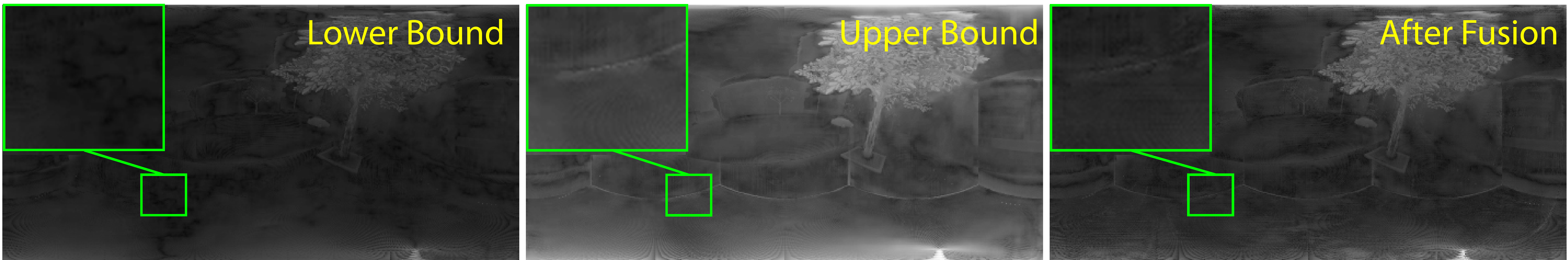}
   \caption{\zfl{Error maps. Green boxes show some fixed errors.}}
   \label{fig:errorbounds}
\end{figure}

The fusion module takes two branches of different projection results. For convenience when training the networks, we firstly trained models for equirectangular, cylindrical, and planar cube-padding, respectively. Thus, we can utilize these models to generate their estimation results and then warp all the flow estimations into the equirectangular projection for training the fusion network. The visualized optical flow results are shown in Fig. \ref{fig:results} and our supplementary materials. \zfl{To quantitatively measure the quality of 360\degree\ optical flows, EPE may not be the best metric, especially for those high-altitude points. We added another metric spherical distance (SD) that
is the geodesic distance on a sphere (also known as great-circle
or orthodromic distance) between a point’s ground
truth and predicted optical flow. In Tab.~\ref{tab:compare_projs}, we report the SDs and EPEs of all the projection methods and their fusion models.} Optical flow results of the models trained by the data with individual projections are all lower than the fusion model where they are involved. Note that equirectangular projection is the best single projection method, given its global continuity and subtle distortions for regions close to the equator. The EPE is 13.7\% lower than the tri-cylindrical projection and 3.1\% lower than the cube-padding projection. The best model is the fusion model from equirectangular and tri-cylindrical projections, with a 5.7\% lower EPE than the second-best model (E+P). The reason is that the cylindrical projection on the cylinders aligned with the z and x-axis can provide more complementary information to the equirectangular projection to improve the prediction quality further. We can also see that even without the best single equirectangular projection, the model trained by fusing tri-cylindrical and cube-padding projection can also get a significantly better result than equirectangular with a 26.6\% lower EPE, demonstrating that our fusion strategy is effective. 
Note that \emph{fused models always outperform} the associated single projection models, which indicates the importance of our fusion principle. \zfl{To further validate the effectiveness of our fusion model, we calculate the lower/upper bound for fusion using an oracle to choose the worst/best flow rather than learning fusion for all our test data. We show in Tab.~\ref{tab:errorbounds} and Fig.~\ref{fig:errorbounds} the lower/upper bound for fusion and ours is closer to the lower bound, \csb{so fusion performs well, but further improvements are possible.}}

\zfl{In our multi-projection process, equirectangular projection
contains the entire scene and provides contextual information
that can be learned by our PWC-based network.
Although the other two projections change the layout of the
scene, we maintain useful local contexts for objects by duplicating
some content. From our results, we can see that
for any given object, PWC’s context network can still gather useful contextual information for that object since most pixels in the surrounding region are not
crossing atlas chart boundaries. Also, the coarse-to-fine scheme of PWC works as long as there is relevant information in the local contextual area in the receptive field of the pyramid. Equirectangular projection always incorporates a full context in a coarse-to-fine scheme due to cyclic convolution operations. For the other two projections, PWC’s context network with six levels of
pyramids has a receptive field that is still relatively local
and does not consider the entire image compared to our resolution of
1000 pixels wide, and is thus applicable to incorporating contextual information.}
\begin{table}[t]
\centering
\begin{tabular}{|l|ll|ll|ll|}
\hline
                             & \multicolumn{2}{l|}{E+C}                   & \multicolumn{2}{l|}{E+P}                   & \multicolumn{2}{l|}{C+P}   \\ \cline{2-7} 
\multicolumn{1}{|l|}{SD/EPE} & \multicolumn{1}{|l|}{CITY100}  & EFT100    & \multicolumn{1}{l|}{CITY100}   & EFT100    & \multicolumn{1}{l|}{CITY100}   & EFT100    \\ \hline
Lower Bound                  & \multicolumn{1}{l|}{0.62/1.31} & 1.83/3.16 & \multicolumn{1}{l|}{0.58/1.44} & 1.88/3.50 & \multicolumn{1}{l|}{0.68/1.68} & 1.88/3.34 \\ \hline
Upper Bound                  & \multicolumn{1}{l|}{1.62/3.74} & 3.96/8.69 & \multicolumn{1}{l|}{1.23/4.29} & 4.08/9.40 & \multicolumn{1}{l|}{1.58/4.27} & 3.91/8.10 \\ \hline
Fusion Error                 & \multicolumn{1}{l|}{0.83/1.79} & 2.64/5.01 & \multicolumn{1}{l|}{0.75/1.95} & 2.95/5.84 & \multicolumn{1}{l|}{0.82/2.15} & 2.77/5.06 \\ \hline
\end{tabular}
\caption{\zfl{Error bounds and the errors of our fused results (SD/EPE).}}
\label{tab:errorbounds}
\end{table}

\subsection{Comparisons}
\textbf{Compare with 2D optical flow estimation methods} We compare our method with the state-of-the-art optical flow estimation methods, FlowNet2, PWC-Net, RAFT, and the method delicately designed for 360\degree\  optical flow estimation \cite{yuan2021360}. \zfl{We feed equirectangular frames to all the compared methods}. The results are shown in Tab.~\ref{tab:compare_others} and Fig.~\ref{fig:results}. Here, the FlowNet2, PWC, and RAFT methods are all using their original method and model to demonstrate the benefits from both of our dataset and multi-projection fusion approach. \zfl{The SD/EPE of our method is \textbf{65.8\%/73.3\%}, \textbf{89.3\%/86.7\%}, and \textbf{54.6\%/73.5\%} lower than PWC, RAFT, FlowNet2 respectively.} 

\noindent\textbf{Comparison with tangent image-based approach~\cite{yuan2021360}} Although the tangent image-based method (TanImg) achieved a large margin of improvement on the panoramic optical flow quality in their own dataset of static objects with moving cameras, \zfl{our SD and EPE are \textbf{41.1\%} and \textbf{42.1\%} lower than TanImg, respectively, as in Tab.~\ref{tab:compare_others}}. This shows that their method cannot adapt to the scenes with dynamic objects because \csb{they do not use a} deep model to learn the useful cues for matching local textures of complex moving objects. We also choose the ``circle" set of data in \cite{yuan2021360} to finetune our model since it is the only set with more than 600 frames (the other two sets both have only 180 frames). We use a 2:1 training and testing split. On the test set, we have an EPE of 4.73 if we directly use the original frames for training and an EPE of 3.8 if we use data augmentation using RGBShift and ChannelShuffle from Albumentations~\cite{info11020125} to create a larger diversity. The result from our data-augmented model is slightly worse than their method (EPE: 3.5) when training by such a small scale dataset. But the significant improvement with only color-based data augmentation indicates that the amount of data is \emph{not sufficient} for training a deep model, and it has a great potential to improve given more data.

\begin{table}[t]
\begin{center}

\begin{tabular}{|p{48pt}<{\centering}||p{50pt}<{\centering}|p{50pt}<{\centering}|p{50pt}<{\centering}|p{50pt}<{\centering}|p{50pt}<{\centering}|}
\hline
SD/EPE   & PWC-Net  & RAFT & TanImg & FlowNet2 & Ours \\ \hline\hline
City100 & 3.86/9.84     & 12.10/21.57 &  1.27/3.69  &2.72/10.85& \textbf{0.82/1.79}     \\ \hline
EFT100  & 6.26/15.64    & 20.21/29.64 &  4.61/8.06  &4.91/14.88& \textbf{2.64/5.01}     \\ \hline
Average &5.06/12.74  &16.15/25.61 & 2.94/5.88 & 3.81/12.87 & \textbf{1.73/3.40}      \\ \hline
\end{tabular}
\end{center}
\caption{\zfl{Comparison with other methods using Spherical Distance (SD) / EPE.}}
\label{tab:compare_others}
\end{table}

\begin{figure}[t]
    \centering
    \includegraphics[width=.95\textwidth]{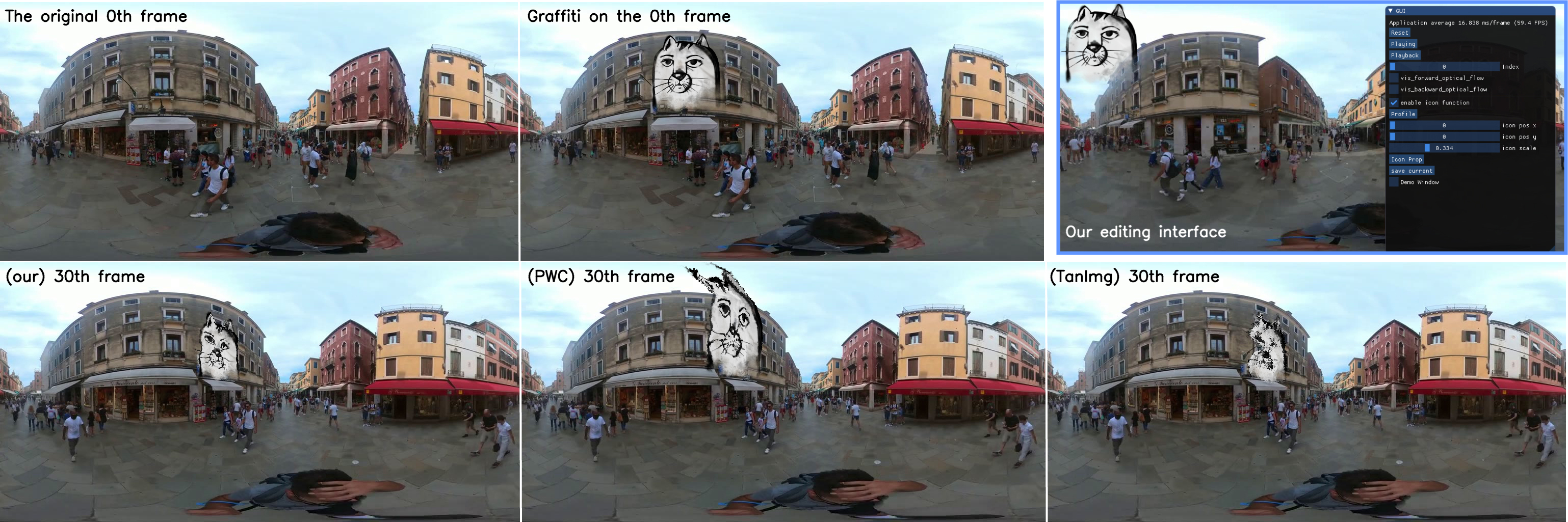}
    \caption{Our video editing application on real-world 360\degree\ videos. The results are generated using optical flow results of our method, PWC, and TanImg respectively. }
    \label{fig:app}
\end{figure}

\subsection{Application for Video Editing}
\zfl{An accurate optical flow estimation is important for many tasks such as object tracking and video segmentation on 360° videos. To demonstrate the accuracy and the practical value of our method for real-world panoramic videos, we develop an application that utilizes the predicted optical flow to propagate edits across frames to evaluate our method visually.} 
In our example shown in Fig. \ref{fig:app}, we put a piece of graffiti at one frame and applied optical flow estimation to find the corresponding affected pixels on the rest of the video frames. More specifically, we use the backward optical-flow to find whether a pixel in the following frames is affected, so that the edits can be propagated densely. We use different optical flow predictions from the PWC model, tangent image-based method, and our method for comparisons. The optical flow estimated by PWC fails to match the distorted corresponding pixels in the upper region. TanImg method is better, but it accumulates errors when the frame number is increasing, leading to obvious artifacts after 30 frames. In contrast, our method well adapts to the distortions of panoramic video representation, generating much more consistent editing results across all the frames. See our supplementary for the full video.

\noindent\textbf{Limitation and Future Work} Our dataset does not contain the real-world optical flow data. In the future, the 360\degree\ Lidar can be used for ground truth optical flow generation.  \zfl{We also consider adding modules to cope with occlusion regions} and extending the work to 360\degree\  scene flow estimation for VR applications.  

\section{Conclusion}
This paper presented a novel multi-projection fusion framework for 360\degree\ optical flow estimation. Besides the commonly used equirectangular projection, we proposed tri-cylindrical projection and circular cube-padding projection to provide complementary less distorted 2D panoramic representations to train the deep model. A fusion network is then applied to combine the warped frames based on the predicted panoramic optical flow using different projections for the final optical flow estimation. A novel dataset was also generated to enable the training of neural networks for 360\degree\ optical flow estimation. The experimental results demonstrate that our method has the best performance and is sufficiently practical to be applied to real-world use cases.

\ \par

\noindent\textbf{Acknowledgements} This work is supported by Marsden Fund Council managed by Royal Society of New Zealand (No. MFP-20-VUW-180). 

\par\vfill\par

\clearpage
%
%
\bibliographystyle{splncs04}
\bibliography{egbib}
\end{document}